\documentclass[10pt,twocolumn,letterpaper]{article}

\usepackage{cvpr}
\usepackage[table, svgnames]{xcolor}
\usepackage{times}
\usepackage{epsfig}
\usepackage{graphicx}
\usepackage{amsmath}
\usepackage{amssymb}
\usepackage{siunitx}
\usepackage{multirow}

\usepackage[acronyms]{glossaries}
\usepackage{tabularx,booktabs}
\usepackage{comment}

\makeglossaries

\newacronym{NLP}{NLP}{Natural Language Processing}
\newacronym{DNN}{DNN}{Deep Neural Networks}
\newacronym{IPU}{IPU}{Intelligence Processing Unit}
\newacronym{ONNX}{ONNX}{Open Neural Network eXchange format}
\newacronym{PopART}{PopART}{Poplar Advanced RunTime}

\newacronym{FP16}{FP16}{float 16 bits}
\newacronym{FP32}{FP32}{float 32 bits}

\usepackage[breaklinks=true,bookmarks=false]{hyperref}

\cvprfinalcopy 

\def\cvprPaperID{****} 

\begin{document}

\title{Graphcore C2 Card performance for image-based deep learning application: A Report}

\author{Ilyes Kacher\\
Qwant\\
\and
Maxime Portaz\\
Qwant\\
\and
Hicham Randrianarivo\\
Qwant\\
\and
Sylvain Peyronnet\\
Qwant\\
}

\maketitle

\begin{abstract}
Recently, Graphcore has introduced an IPU Processor for accelerating machine learning applications. The architecture of the processor has been designed to achieve state of the art performance on current machine intelligence models for both training and inference.

In this paper, we report on a benchmark in which we have evaluated the performance of IPU processors on deep neural networks for inference. We focus on deep vision models such as ResNeXt. We report the observed latency, throughput and energy efficiency. 
\end{abstract}

\section{Introduction}
\label{introduction}
\gls{DNN} approaches are being increasingly deployed into real-time applications across a wide spectrum of functional domains,  ranging, for example, from video segmentation to natural language translation.

The real-time nature of these applications creates a  requirement for system responsiveness, which imposes strict latency constraints on the inference of the large underlying  \gls{DNN} models.

A response latency of 100 ms or less to a query sent by a user will give the impression of instantaneous system feedback. The latency limit for uninterrupted utilization of an interactive system is about one second. When the response latency reaches two seconds or more, the user will likely doubt that the system is working properly or will switch attention to another task altogether~\cite{nielsen1994usability}. Having the smallest response time is thus crucial for industrial real-time application of machine learning.

A more efficient use of hardware can improve the performance and scalability of an application without incurring additional cost, which is critical in massively-deployed consumer applications.

One example is a batching strategy that consists of aggregating multiple inputs into a single batch load. However, real-time latency limits the batch size~\cite{amodei2016deep} as the latency increases proportionally with size. Furthermore, uneven system loading will limit the possibility of building up optimal batch sizes, as the waiting time for the queue to fill with the right amount of inputs is constrained by the overall response latency.

At Qwant, we have developed a prototype of an image search engine\footnote{\url{https://research.qwant.com/images/}} \cite{portaz2019image,portaz2020demo} based on \gls{DNN} and are motivated to optimise performance for both training and inference.  

Graphcore recently introduced the \gls{IPU} processor, developed to accelerate machine learning applications. The IPU is a processor designed for parallel computation of sparse high dimensional graphs and data structures. It supports massively parallel processing across thousands of independent
processing threads. This is achieved by the 1,216 high performance machine learning processor cores (IPU-Cores) on the IPU, each of which contains 6 processor threads. Memory is distributed on the chip. Each IPU-Core is coupled to 256kB of memory, yielding 304MB of SRAM memory per IPU, and a memory bandwidth of 45TBps. The IPU adopts a Bulk Synchronous Parallel (BSP) approach to facilitate efficient programming~\cite{jia2019dissecting}.

The Graphcore C2 card is a PCI Express Gen3/4 card containing two IPUs. One C2 card draws equivalent power (300W) to alternative single chip offerings on the market.

The IPU processor can be programmed with Graphcore's Poplar SDK. We used PopART (Poplar Advanced Runtime), a graph runtime that takes computational graphs, currently ONNX models, and provides IPU-specific optimisations for inference and training. Users can also run graphs described by machine learning frameworks TensorFlow and PyTorch via the Poplar SDK.

We report in this paper on the benchmark we performed on the \gls{IPU} Processor. We focus on evaluating the performance for inference tasks.

We evaluate latency, throughput, and energy efficiency, as good indicators of hardware performance~\cite{PLASTER}. Latency is the time between the user request and response. Throughput is the number of inference tasks per second that the system can complete. Energy efficiency is the inference task's total power consumption for the whole system. Throughput is measured in images per second for image-based \gls{DNN}, and energy efficiency measured in images per second per Watt.

This paper evaluates C2 card performance on an image-based neural network in terms of latency, throughput, and energy efficiency.

\subsection{Overview}
\label{sec:overview}

This report summarizes the performance of one C2 card performing inference on the recent image-based deep learning model ResNeXt101~\cite{xie2017aggregated}.

For the evaluation, we make use of the full compute capacity of the C2 card by using its two \gls{IPU} processors~\footnote{\url{https://www.graphcore.ai/technology}} with each  processor  running  one  inference  session  in  parallel. Access to \gls{IPU} technology was delivered through the Microsoft Azure cloud (Preview Access to NDv3 Graphcore IPU powered Azure VM instances~\cite{msft})

The implementation uses a PyTorch model which is exported to the industry standard \gls{ONNX} to run in PopART~\footnote{\url{https://www.graphcore.ai/posts/graph-computing-for-machine-intelligence-with-poplar}}.

As detailed in section~\ref{sec:implementation}, the performance metrics of the evaluation are:
\begin{itemize}
 \item Latency: time per batch,
 \item Throughput: number of images per second,
 \item Energy efficiency: number of images per second per Watt.
\end{itemize}

We use these metrics to evaluate the performance of the C2 card for a real-time application and show that the IPUs provide a lowest latency of $1.36$ ms and highest throughput of $2526.35$ images per second, depending on the batch size. Details are given in section \ref{sec:results}.

\section{Background and related work}
\label{sec:background_related_work}


The use of neural networks has increased with many applications: classification, segmentation and language processing.

ResNeXt101~\cite{xie2017aggregated} with its 44M of parameters stored in 176 MB as 32-bit numbers is an image-based example of a large model.

For a typical application involving machine learning, 90\% of the production cost is spent on inference~\cite{aws_keynote} tasks. 
In real-time application, the latency is bound to the computation time during inference. Conversely, the training step is a lengthy phase that will affect the quality of the result but that does not impact user latency. 

A simple way to increase the response rate of an application is to gather multiple queries into a single batch for the model inference. However, there is a trade-off between the response time and the batch size. Large batch sizes benefit from high parallelism but increase user latency. ~\cite{amodei2016deep} studies the effect of batching inputs on 10 to 30 concurrent users. Their study shows that 90\% of the queries are treated in a batch of size 4 with a limit of 10.

In their work, S. Gupta et al.~\cite{gupta2015deep} show that training a \gls{DNN} using mixed-precision number representation with stochastic rounding results in little to no degradation in the model performance. Mixed-precision accelerates training and inference. This is especially true on hardware with specialized mixed-precision on chip, such as the Graphcore C2 card. 


\section{Benchmarks}

\subsection{Experimental setup}

\subsubsection{Hardware}
Our experiments use one C2 card with its two IPU processors.

\begin{figure*}[!h]
    \centering
    \includegraphics[width=\textwidth]{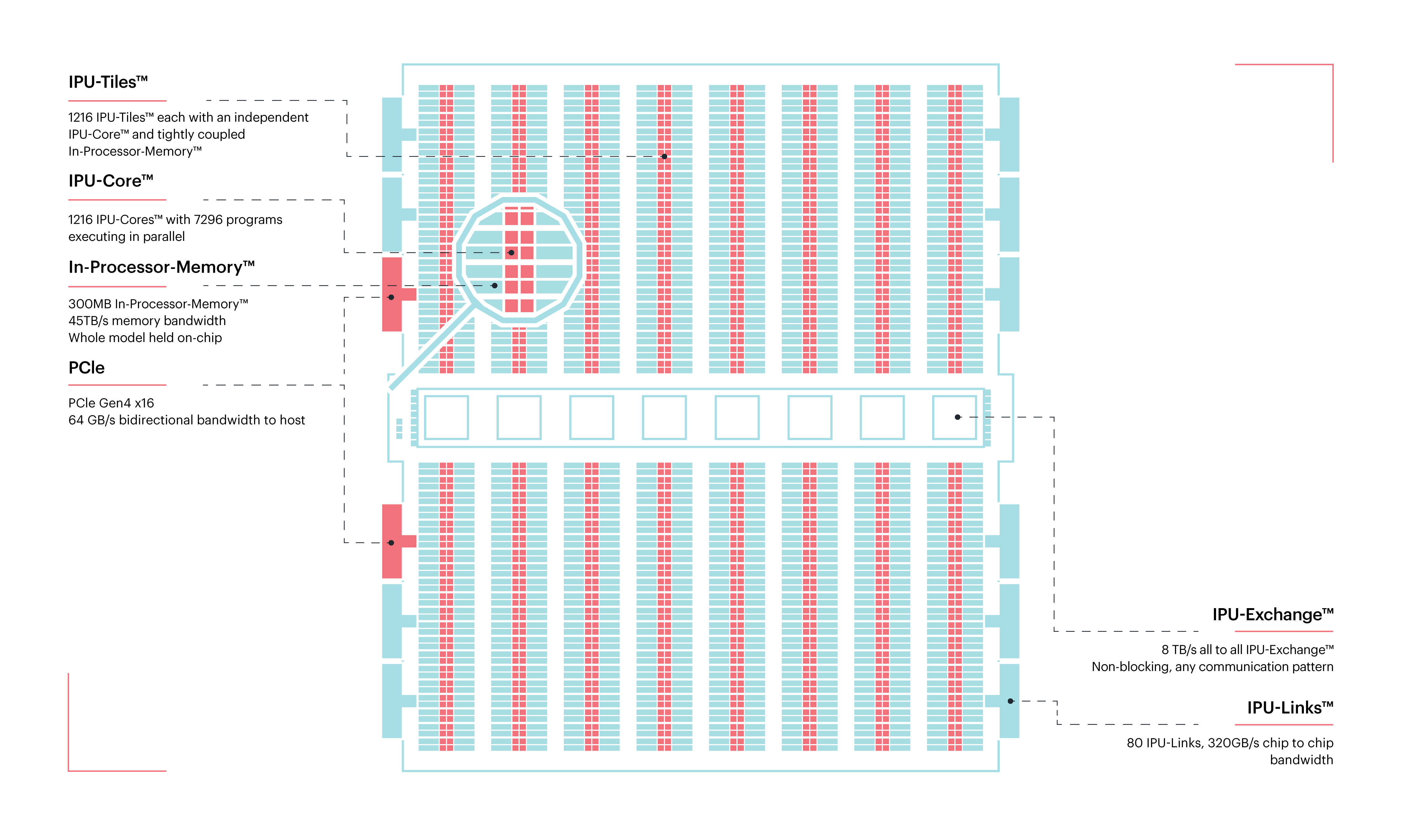}
    \caption{IPU processor diagram -- republished with permission from Graphcore.}
    \label{fig:latency}
\end{figure*}

\subsubsection{Implementation}
\label{sec:implementation}

Our implementation uses the PopART (Poplar Advanced Runtime) library provided with the Poplar SDK~\cite{poplar}. We used SDK version 1.0.136.

We import a ResNeXt101 model from PyTorch~\cite{cadene} format to \gls{ONNX}, to run in PopART in mixed precision. Our implementation acquires one IPU device to perform the inference on the model. The images used for the inference are a subset of 10,000 images from the COCO validation set~\cite{lin2014microsoft}. One instance of the implementation is launched on each IPU of the C2 card.

We measure the computation time of each running instance alongside the C2 card power consumption. We use \texttt{gc-monitor} from Graphcore driver utilities to measure the power consumption of one C2 card. 
The measurements are used to compute the latency, throughput and energy efficiency of the card.

\textbf{Latency} is measured in milliseconds per batch. It represents the overall time to get the output results from an input batch. The latency contains the following operation: preprocessing of the input batch, inference, and retrieval of the output from the device.

\textbf{Throughput} is measured in images per second and is obtained from the latency time measurement and the batch size. This measure represents the load that the hardware can handle for an image-based deep learning application.

\textbf{Energy efficiency} is measured in images per second per Watt. It represents the energy effectiveness of the hardware on an image-based deep learning application. The power is measured multiple times over a few minutes of inference and averaged. The energy efficiency is the throughput divided by the average power.

\subsection{Results}
\label{sec:results}
In tables \ref{tab:latencyIPU}, \ref{tab:throughputIPU}, \ref{tab:efficiencyIPU} we present the latency, throughput, and energy efficiency results of the C2 card for ResNeXt101~\cite{xie2017aggregated} inference tasks. In the experiment, the batch size corresponds to the number of input given to the C2 card. The micro batch size refers to the input of one IPU processor. We experiment with batch size of {2, 4, 6, 8, 10, 12} on the C2 Card which translate to a micro batch size {1, 2, 3, 4, 5, 6} per IPU processor.

The C2 card shows the lowest latency of $1.36$ ms on batch size 2 and the highest latency of $4.75$ ms on batch size 12. The best throughput obtained is $2526.35$ images per second with a batch size of 12. The C2 card is more efficient energy-wise on a larger batch size with $9.68$ images per second per watt on batch size 10. 

Overall the C2 card has excellent latency for real-time applications with high throughput capabilities. It reaches a stable images per second to power ratio from batch size 8. This makes batch sizes 8 to 12 good candidates to choose for optimal latency or throughput depending on the application without affecting the energy cost too much.

We tested batch sizes up to 12 (6 per IPU processor) on the C2 card. For large batch sizes, Graphcore's hardware and SDK support efficient data parallel training and inference over multiple IPUs. While ResNeXt101 fits into the in-processor memory of a single IPU in half precision, larger models can be run in a model parallel manner using pipelining, which can be controlled via the Poplar SDK. 

The benefit of training using a small batch size has been studied in~\cite{masters2018revisiting}. Furthermore, due to latency constraint small batch sizes are mostly used in real-time inference applications~\cite{amodei2016deep}.

\begin{table}[]
 \centering
 \rowcolors{1}{white}{gray!15}
 \begin{tabular}{@{}SS@{}}
 \toprule
 {Batch size} & {Latency} \\
 \midrule
 2 & \bfseries 1.36\\
 4 & 1.94\\
 6 & 2.82\\
 8 & 3.32\\
 10 & 4.13\\
 12 & 4.75\\
 \end{tabular}
 \caption{Latency results for ResNeXt101 (milliseconds per batch)}
 \label{tab:latencyIPU}
 \end{table}

\begin{table}[]
 \centering
 \rowcolors{1}{white}{gray!15}
 \begin{tabular}{@{}SS@{}}
 \toprule
 {Batch size} & {Throughput} \\
 \midrule
 2 & 1474.16\\
 4 & 2063.03\\
 6 & 2131.01\\
 8 & 2409.79\\
 10 & 2421.82\\
 12 & \bfseries2526.35\\
 \end{tabular}
 \caption{Throughput results (images/second) for ResNeXt101}
 \label{tab:throughputIPU}
 \end{table}
 
\begin{table}[]
 \centering
 \rowcolors{1}{white}{gray!15}
 \begin{tabular}{@{}SS@{}}
 \toprule
 {Batch size} & {Energy efficiency} \\
 \midrule
 2 & 6.51\\
 4 & 7.80\\
 6 & 7.87\\
 8 & 9.07\\
 10 & \bfseries9.68\\
 12 & 9.49\\
 \end{tabular}
 \caption{Energy efficiency results (images/second/watt) for ResNeXt101}
 \label{tab:efficiencyIPU}
 \end{table}

\section{Conclusion}
The C2 card provides fast and efficient performance for inference tasks. In our experiment, we report the lowest latency of $1.36$ ms and the highest throughput of $2526.35$ images per second on ResNeXt101 (respectively for batch size 2 and 12). 

The C2 card is a promising technology that deep learning practitioners should keep an eye on. 
Worthwhile future studies to complement this paper include the evaluation of pipelining large models over multiple IPUs and the analysis of training performance.


\end{document}